\documentclass[conference]{IEEEtran}
\usepackage{times}
\usepackage[pdftex]{graphicx}
\graphicspath{{./}}
\DeclareGraphicsExtensions{.pdf,.jpeg,.jpg,.png}
\usepackage[caption=false,font=footnotesize]{subfig}

\usepackage[numbers]{natbib}
\usepackage{multicol}
\usepackage[bookmarks=true]{hyperref}
\usepackage{hyperref}
\usepackage{mathptmx} 

\begin{document}

\title{Deep Reinforcement learning for real autonomous mobile robot navigation in indoor environments}


\author{\authorblockN{Hartmut Surmann, Christian Jestel, Robin Marchel, Franziska Musberg, Houssem Elhadj and Mahbube Ardani}
  \authorblockA{Computer Science Department, University of Applied Science Gelsenkirchen, \\
    Neidenburgerstr. 43, D-45877 Gelsenkirchen, Germany, Email: hartmut.surmann@w-hs.de}}


%

\maketitle

\begin{abstract}
Deep Reinforcement Learning has been successfully applied in various computer games \cite{DBLP:journals/corr/MnihKSGAWR13}. 
However, it is still rarely used in real-world applications, especially for the navigation and continuous control of real mobile robots \cite{DBLP:journals/corr/abs-1804-10332}. 
Previous approaches lack safety and robustness and/or need a structured environment. 
In this paper we present our proof of concept for autonomous self-learning robot navigation in an unknown environment for a real robot without a map or planner. 
The input for the robot is only the fused data from a 2D laser scanner and a RGB-D camera as well as the orientation to the goal. 
The map of the environment is unknown. 
The output actions of an Asynchronous Advantage Actor-Critic network (GA3C) are the linear and angular velocities for the robot. 
The navigator/controller network is pretrained in a high-speed, parallel, and self-implemented simulation environment to speed up the learning process and then deployed to the real robot. 
To avoid overfitting, we train relatively small networks, and we add random Gaussian noise to the input laser data. 
The sensor data fusion with the RGB-D camera allows the robot to navigate in real environments with real 3D obstacle avoidance and without the need to fit the environment to the sensory capabilities of the robot. To further increase the robustness, we train on environments of varying difficulties and run 32 training instances simultaneously.
\footnote{Video: supplementary File / YouTube, Code: GitHub}
\end{abstract}

\IEEEpeerreviewmaketitle

\section{Introduction}
Since the 90s of the last century, one of the major challenges for robot control and navigation is safe and robust collision avoidance, so that the robot can navigate from its starting position to its goal position \cite{7881744}. 
This holds for user programmed robots as well as for policies learned by the robot itself to control and navigate. 
In the area of autonomous robots, learning strategies, especially reinforcement and deep learning, gained much interest in recent years. 
Unfortunately, most of the robot learning approaches use only simulations or artificial environments and are not transferred to real robots in real (not controlled) environments. 
The challenges to achieve that, especially for reinforcement deep learning based robots, are: 
\begin{enumerate} 
\item {\bf Simulation}: Since reinforcement learning tests millions of input/output combinations, we need a fast (parallel) simulation environment.
Another requirement for the simulator is to model the real world accurately and to support a wide range of environments (from simple to complex). 
\item {\bf Robot deployment}: To deploy and to continue the learning/testing on a real robot we need (besides the robot) a software framework (i.e., ROS) and a strategy to transfer the learning results from the simulation to the robot and its sensors. The different sensors should be fused (sensor fusion).  
\item {\bf The Network}: The deep reinforcement learning system should observe the speed and memory constraints introduced by the robot platform. Therefore, a Network architecture, size, number of parameters, learning strategy, as well as input and output parameters, have to be found.  
\end{enumerate}

\begin{figure}[!t] 
\centering 
\includegraphics[width=0.48\textwidth]{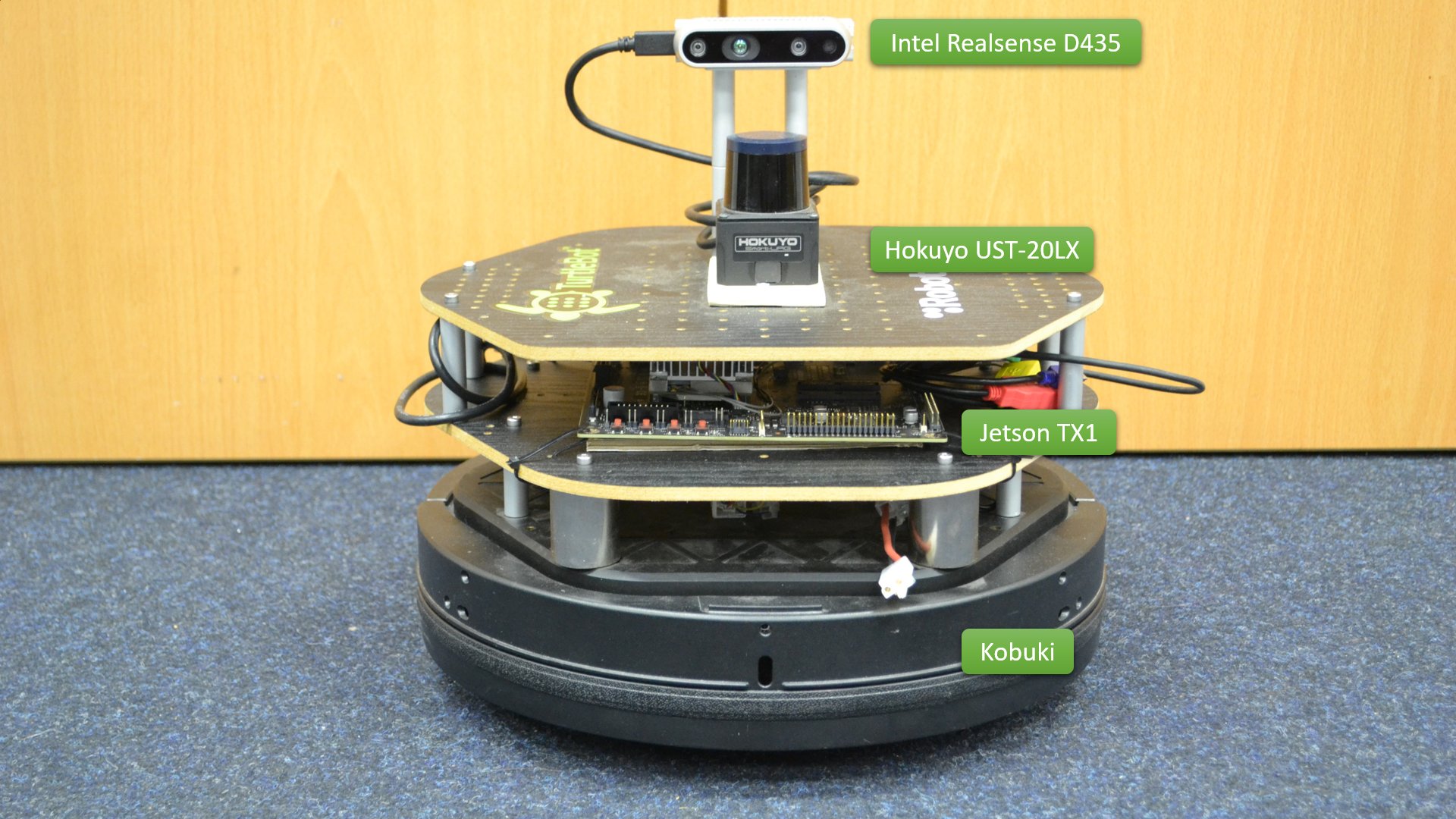} 
\caption{The robot consists of a turtlebot 2 (Kobuki) with a Nvidia Jetson TX1, a Hokuyo UST-20LX and an Intel Realsense D435.} 
\label{fig:robot} 
\end{figure} 

\section{Related work for Deep-Learning-based navigation} 
Shabbir et al. \cite{DBLP:journals/corr/abs-1803-07608} give an in-depth research survey of deep learning techniques for mobile robot applications, with a specific focus on the advantages and obstacles, in comparison to conventional robotics.
They show that the focus of current deep learning research is vision-based, whereas most current real mobile robotic applications are based on laser scanners.    
Chen et al. \cite{DBLP:journals/corr/ChenELH17} build a robot with a 3D laser scanner and a deep learning network with a navigation policy that respects common social norms and can avoid obstacles. The path planner, however, does not follow a deep learning approach.
Lillicrap et al. \cite{DBLP:journals/corr/LillicrapHPHETS15} presented an actor-critic, model-free algorithm, based on the deterministic policy gradient that can operate over continuous action space, but only apply it to a simulated environment. 
Tai et al. \cite{DBLP:journals/corr/TaiPL17} presented a learning-based mapless motion planner by taking the sparse 10-dimensional 
range findings and the target position with respect to the mobile robot coordinate frame as input and the continuous steering commands as output. 
They also deploy their deep learning network to a physical robot platform. 
Since the absence of a 3D sensor, they have to prepare the environment to avoid obstacles outside of the 2D plane sensed by the laser scanner. 

Instead of sparse range findings, our approach uses the full number of 1081 range findings of the scanner fused with the 3D data of an RGB-D camera. 
Since available robotic simulation frameworks (e.g., Gazebo) are to slow for neural network learning purposes, we implemented a fast and parallel robot simulation; this enables us to simulate different environments with different complexities in parallel which boosts the learning speed of the network. 
The outline of the paper is as follows. In section \ref{robot} we describe the robot and its sensors followed by 
section \ref{simulation} and section \ref{network} where we describe the implemented simulation environment and the inner workings of the deep learning network. Section \ref{training} describes the training and evaluation before section \ref{sec:conclusion} concludes the paper.

\section{The Robot} 
\label{robot} 
The hardware forms the central unit of the robot platform (Fig. \ref{fig:robot}) and the neural network. 
With the sensors, the robot platform can perceive the environment and interact with it through its actuators. 
For a simple integration of the whole system, it was essential to ensure that the hardware and sensors are compatible with ROS. 
The base of the robot platform is the Turtlebot 2 (Kobuki, 354 x 354 x 420 mm). 
It is a robot that is easy to configure, and with its simple plate design, the hardware and the sensors can be placed arbitrarily. 
Due to its ROS compatibility, the Turtlebot can be easily controlled. 
It can reach a linear velocity of 0.65 m/s, an angular velocity of 180$^\circ$/s and can carry a payload of up to 5 kg.

\begin{figure}[!t] 
\centering 
\subfloat[UAVs]{\includegraphics[width=0.48\linewidth]{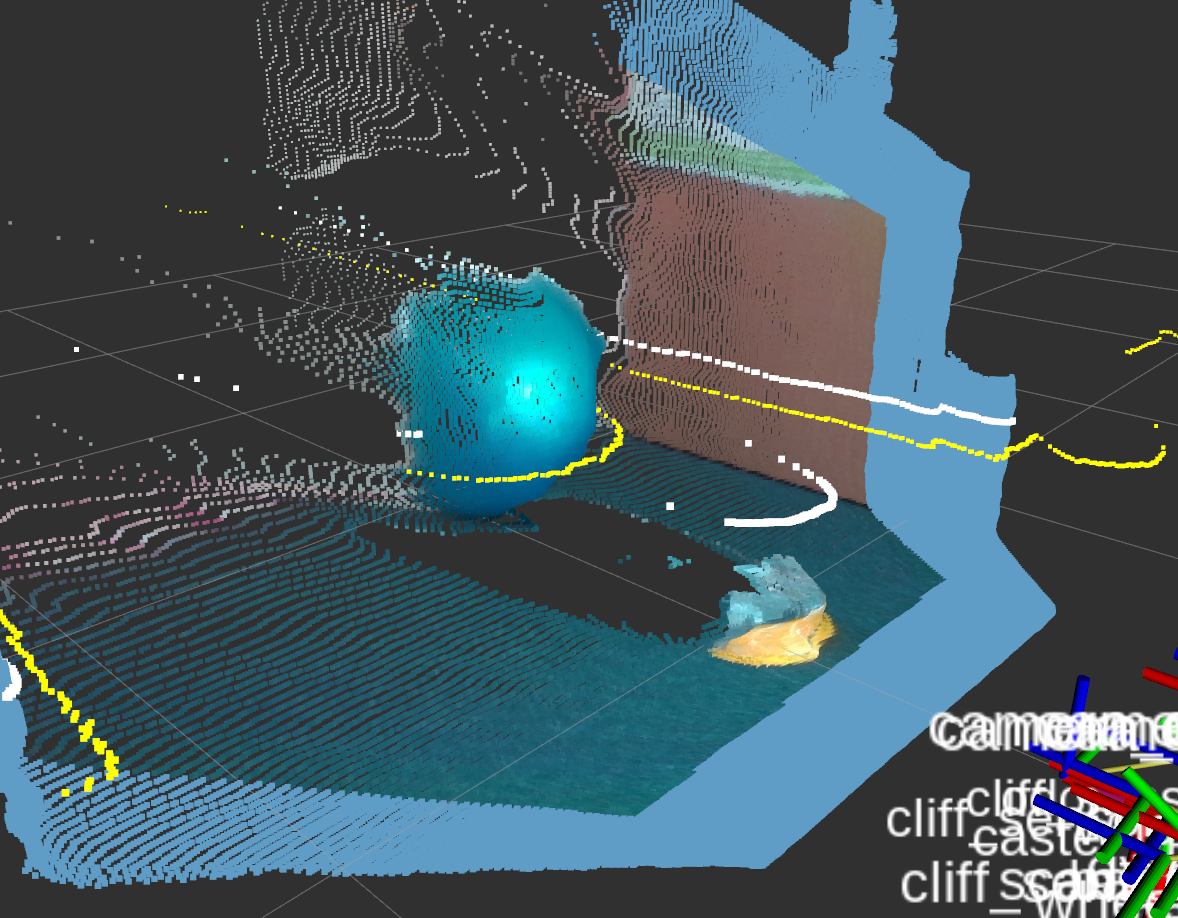}} 
\hfil 
\subfloat[UGV]{\includegraphics[width=0.48\linewidth]{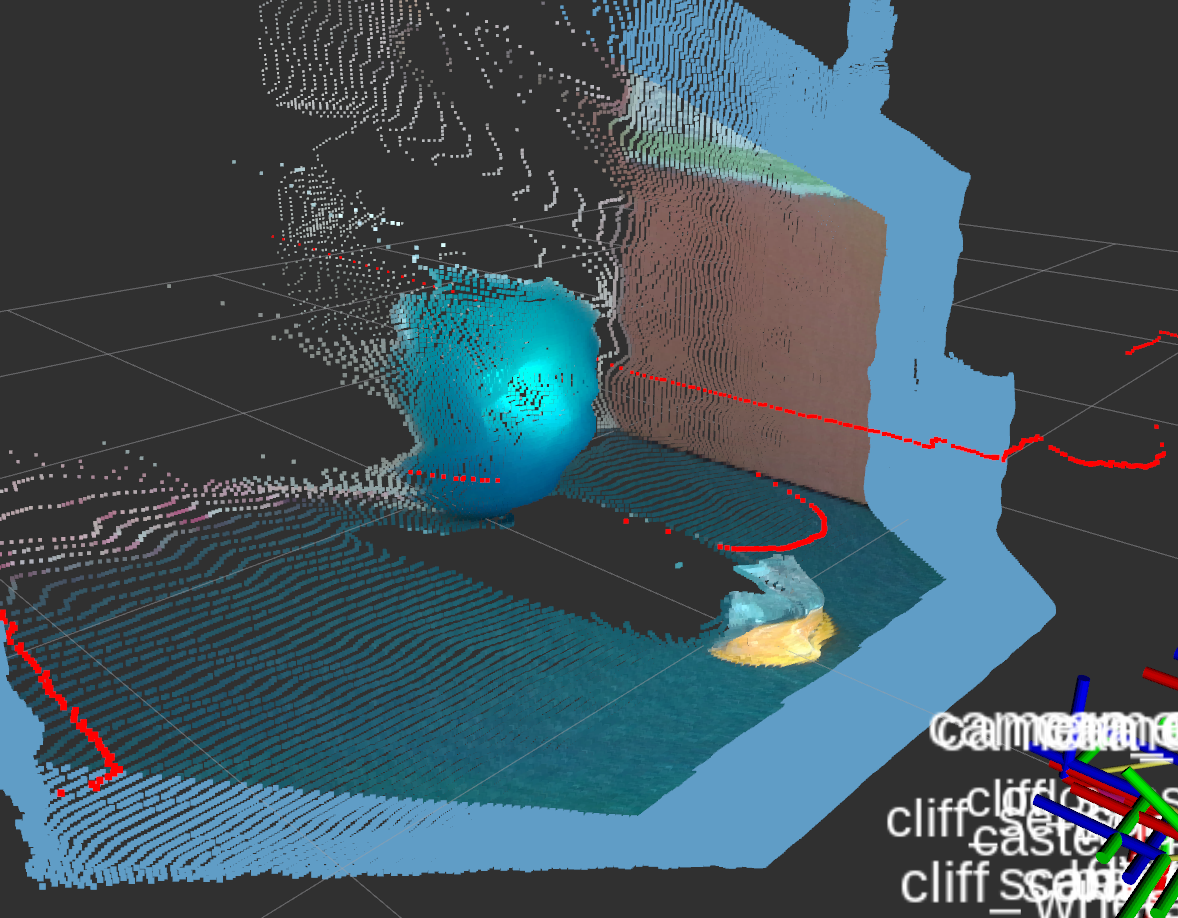}} 
\caption{Left: Hokuyo laser scan (yellow) and the computed 2D laser scan (out of the 3D point cloud) from the Realsense (white). Right: The sensor fusion of both sensors to one 2D laser scan.} 
\label{fig:sensorfusion} 
\end{figure} 

The brain of the robot consists of an Nvida Jetson TX1 board (Quad-core ARM Cortex-A57, 256 core Maxwell GPU, 4GB LPDDR4 RAM, 16GB eMMC storage). 
The board is compact and fits exactly one plate of the Turtlebot 2. 
The required board input voltage is between 5.5-19.6V and comes from a lithium polymer battery which operates the robot for four hours. 
The board has a GPU which is supported by tensorflow 1.7 on cuda 10.

The primary sensors are a Hokuyo UST-20LX and the Intel Realsense D435. 
The Hokuyo UST-20LX (range: 0.06 - 20m, angle: 270$^\circ$, resolution: 0.25$^\circ$, precision: +/- 40 mm, frame rate: 40 Hz) is a compact 2D laser scanner. 
With its opening angle of 270$^\circ$, the robot can almost perceive its entire surroundings and recognize sharp turns up to 135$^\circ$. 
The second sensor, the Intel Realsense D435 is an RGB-D sensor. 
It has a horizontal opening angle of approx. 90$^\circ$ which means only one-third of the field of view of the laser scanner is covered. 
In contrast to the laser scanner, the RGB-D sensor has a vertical opening angle of approx. 60$^\circ$, which enables a 3D scan of the surroundings. 
Thus, obstacles over and under the scan height of the laser scanner can be detected. 
Unfortunately, the Realsense has relative strong noise, which increases with the distance. 
Noisy sensor values can lead to ghost obstacles and uncertain behavior of a robot. 
So, compared to the RGB-D sensor, the laser scanner is more accurate.  
The Realsense has a field of view of 86 x 57 x 94$^\circ$ (+/- 3$^\circ$), a resolution up to 1280 x 720, a frame rate of up to 90 fps, distance 0.2 - 10m.

\section{The Simulation} 
\label{simulation} 
\begin{figure}[!t] 
\centering 
\includegraphics[width=0.48\textwidth]{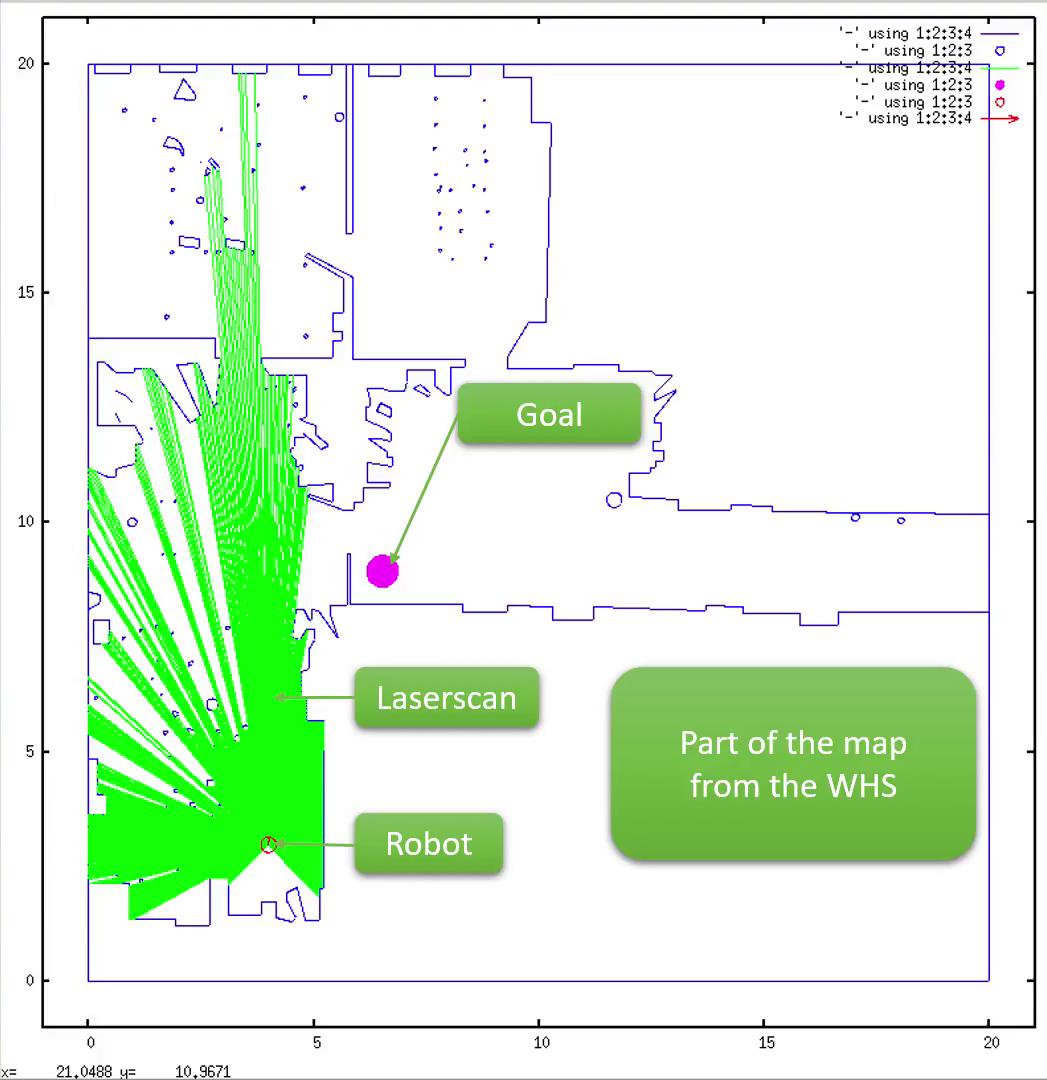} 
\caption{Simulation2d: The map shows the robotics lab of the university campus Gelsenkirchen. The map is built with the ROS package hector slam.  
Target Point (Magenta), Laser Beams (Green), Robot (Red).} 
\label{fig:simulation} 
\end{figure} 
\begin{figure}[!t] 
\centering 
\includegraphics[width=0.48\textwidth]{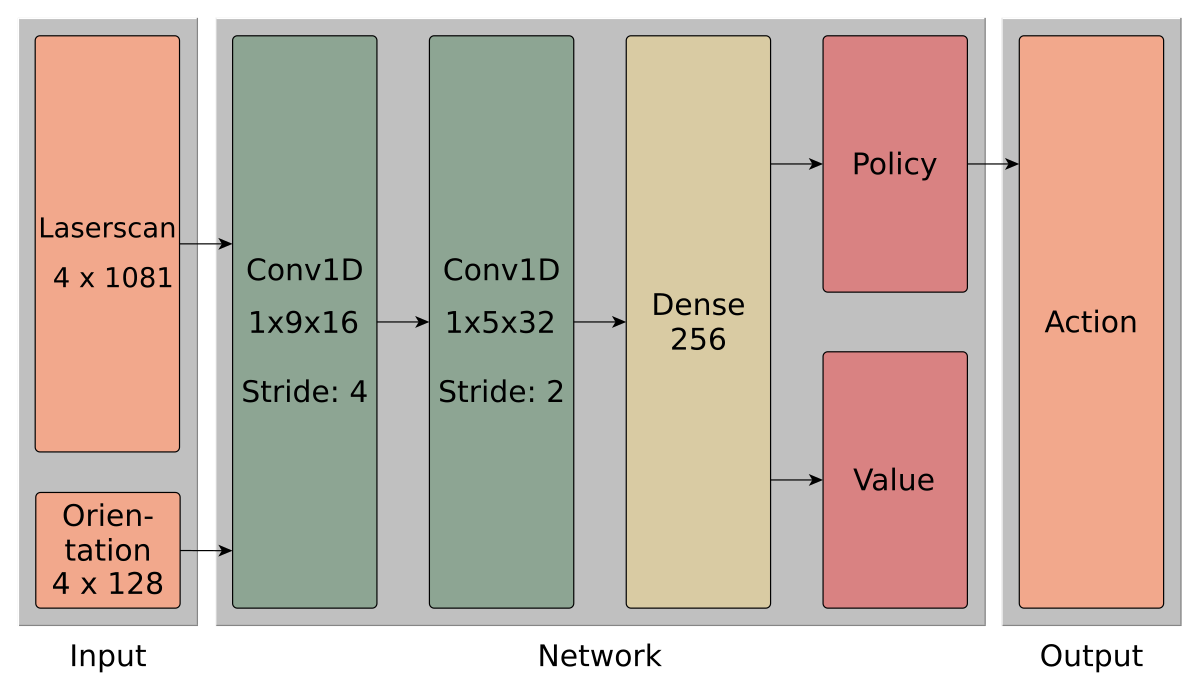} 
\caption{Architecture of the actor critic deep learning network trained with reinforcement learning.} 
\label{fig:architectur} 
\end{figure} 

One crucial step is to fuse the two different sensor values. 
Thus, the advantages of the wide opening angle of the laser scanner and the 3D perception of the RGB-D sensor could be combined (Fig. \ref{fig:sensorfusion}). 
Therefore, the 3D point cloud from the RGB-D sensor is converted to a 2D laser scan (2D fake laser, virtual laser scan). 
The robot operating system ROS\footnote{http://wiki.ros.org} already has a package pointcloud\_to\_laserscan\_node to convert the point cloud to a 2D laser scan. 
The 3D point cloud is filtered according to the height (Z-axis) of the robot, starting from the ground, and transformed into a 2D laser scan according to the horizontal field of view of the RGB-D camera.  
The final fused laser scan is now the pointwise minimum of the 2D laser scan and the converted scan from the RGB-D sensor. 
So, the input vector for the neural network stays small (similar to a 2D laser scan) but with additional 3D information from the environment.

A critical trait of a robot AI is the ability to dream or in other words to simulate its behavior as humans do and to learn from it for real life. 
Popular robot simulation environments like Gazebo (\cite{1389727}), Morse \cite{5980252} or V-REP \cite{6696520} can be used together with ROS. 
The functional behavior of the robot, the laser scan, and the environment can be simulated with all the robot/environment simulators. 
In general, learning environments for neural networks need not only to be functional but also very fast, memory-efficient, and able to run several instances in parallel.  
Therefore, we built a 2D simulation environment (Fig. \ref{fig:simulation}) to train the network and customized it for the robot with its sensors. 
The simulation is written in C++, this allows the use of specific processor instruction sets like AVX for SIMD commands (x86, Intel up Sandy, AMD up Bulldozer). 
The AVX register has a length of 256-bits and can handle eight floating-point operations simultaneously (laser beam intersections). 
This results in an acceleration factor of about seven. 
Obstacles are represented by combinations of circles and lines, and the robot platform is modeled as a simple circular shape. 
The laser scan is at the origin of the robot platform and the properties of the laser scan completely comply with the specifications of the UST-20LX (1081 sensor values, 270$^\circ$). 
We also add Gaussian noise to each laser beam to simulate the measurement error.  
Especially the use of the AVX instruction set for the laser beam intersection of the laser and its environment results in a noticeable speedup. 
For the sake of simplicity, we used the proven plotting tool gnuplot \footnote{http://www.gnuplot.info/} (starts 1986) for the visualization.

Environmental maps can be created in several ways. 
Maps from real environments are built from runs of the real robot, using hector slam with the laser scanner data. \cite{KohlbrecherMeyerStrykKlingaufFlexibleSlamSystem2011}.
Inkscape is used as an editor to create hand-drawn maps, the vector graphic (SVG) is then converted to the simulation format with a python script. Alternatively, maps can be directly generated with python.

\section{The Network} 
\label{network} 
\begin{figure}[!t] 
\centering 
\includegraphics[width=0.48\textwidth]{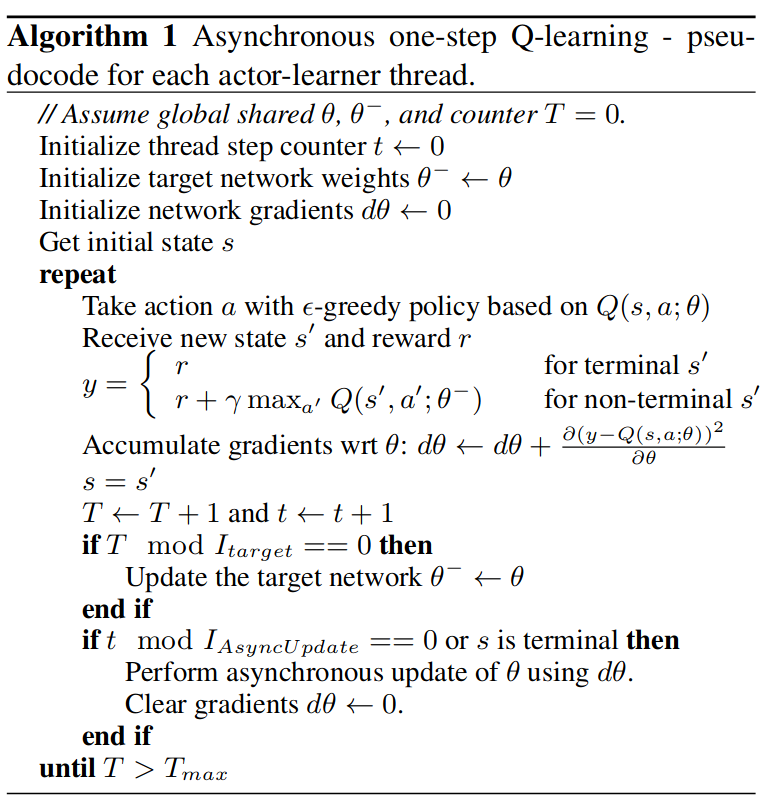} 
\caption{Asynchronous one-step Q-learning - pseudocode for each actor-learner thread \cite{DBLP:journals/corr/MnihBMGLHSK16}.}  
\label{fig:a3c-algo} 
\end{figure}

\begin{figure*}[!t] 
\centering 
\includegraphics[width=0.94\textwidth]{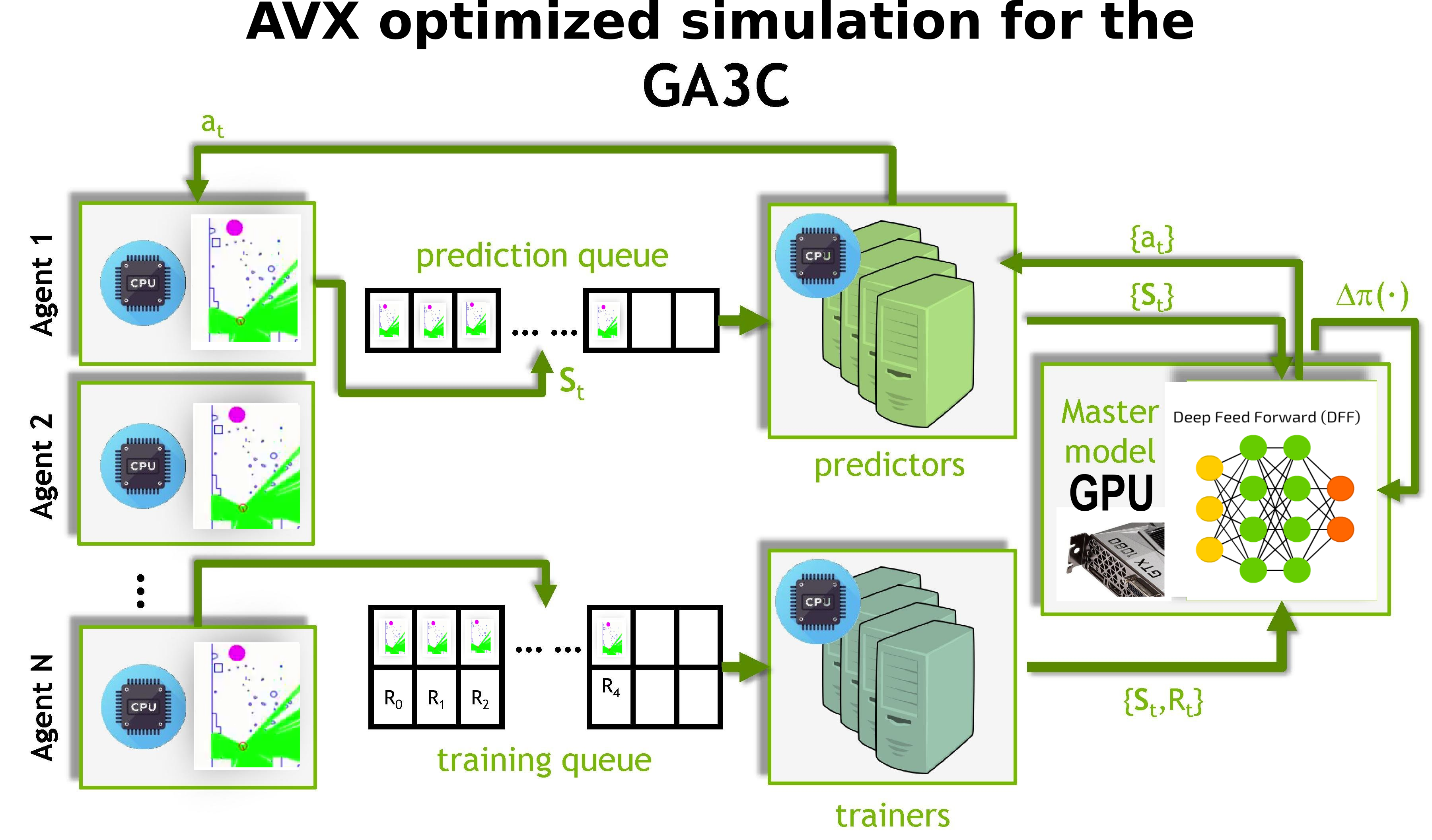} 
\caption{Diagram of parallel learning with GA3C \cite{DBLP:journals/corr/BabaeizadehFTCK16}.}  
\label{fig:ga3c-block} 
\end{figure*}

For training the robot we used the Asynchronous Advantage Actor-Critic algorithm \cite{DBLP:journals/corr/MnihBMGLHSK16} with deep reinforcement learning. 
The Advantage Actor-Critic algorithm has been successfully used to achieve super-human performance in a variety of video games \cite {DBLP:journals/corr/WangBHMMKF16}. 
Babaeizadeh et al. implemented a fast hybrid CPU/GPU version of the Asynchronous Advantage Actor-Critic algorithm (GA3C) \cite{DBLP:journals/corr/BabaeizadehFTCK16} which we used.
Figure \ref{fig:ga3c-block} shows the block diagram of the learning algorithm and 
figure \ref{fig:a3c-algo} the pseudocode for an actor-learner thread of the asynchronous one-step Q-learning.

\begin{figure}[!t] 
\centering 
\includegraphics[width=0.48\textwidth]{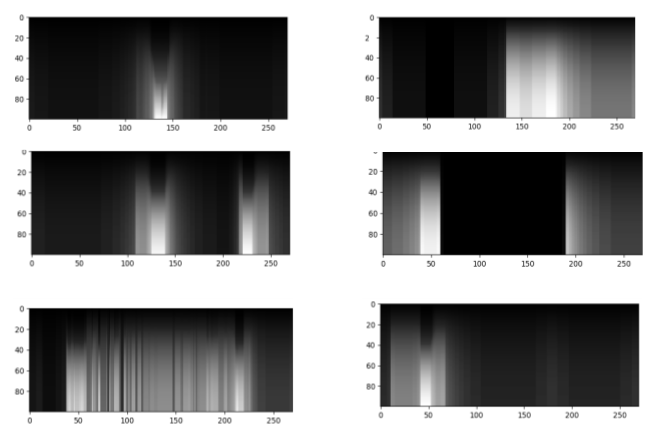} 
\caption{Example patterns of the coding of the laser scanner as an input vector for the GA3C. The laser scans are coded in polar coordinates ($\phi$, $r$) ranging form [0:269] whereas index 0 corresponds to -135 and 269 corresponds to +135. More white means more free space.}  
\label{fig:input-coding} 
\end{figure} 
\begin{figure*}[!t] 
\centering 
\includegraphics[width=0.94\textwidth]{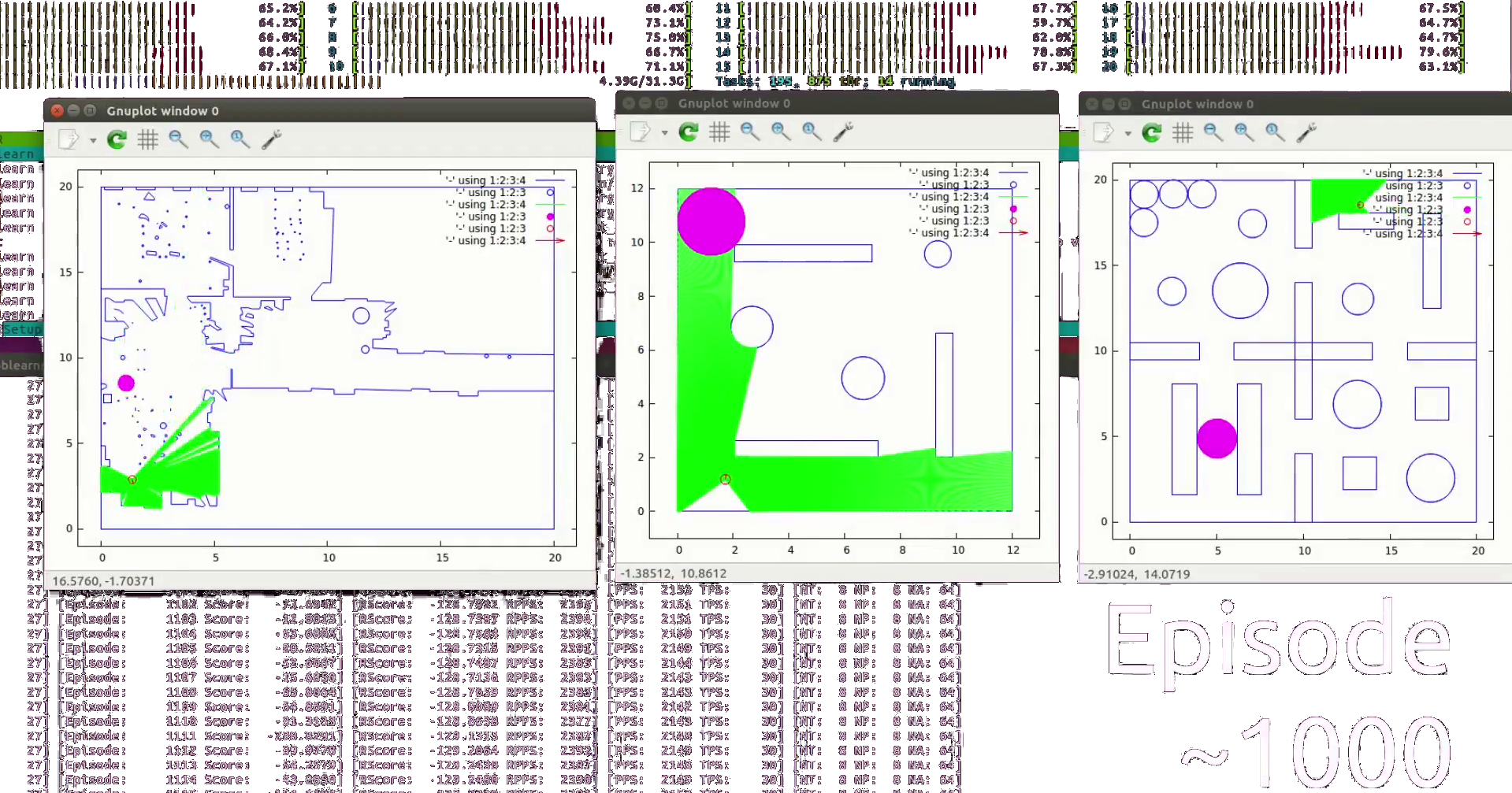} 
\caption{Three parallel instances of the simulation while training } 
\label{fig:simulation3} 
\end{figure*} 

GA3C still uses the massive parallelism of the A3C's agents but each agent no longer has its copy of the global network i.e. only one instance exists. 
Since GA3C uses only one copy of the DNN model, it centralizes predictions and training updates and removes the need for synchronization. 
This network can be readily executed on the GPU since there are no longer several networks around the scarce resource of the GPU. 
The bottleneck, which is the data transfer between CPU / RAM and GPU, is moderated. 
On the other hand, new difficulties arise concerning the communication of the agents with the network. 
New race conditions and problems may arise as a Tensorflow session is not designed for use in multiprocessing environments. 
The GA3C algorithm, therefore, uses two global queues, a prediction queue, and a learning queue. 
In the former, an agent lines up when it needs a prediction for an observation. 
A separate prediction process now handles the processing of the queue and the communication with the global network. 
The process then sends the results back to the individual agents, that can now resume their work. 
If an agent has reached a terminal state or the maximum number of steps it collects the experience gained, 
accumulates the total reward (R) for each step, taking the discount factor into account 
and puts the experiences into the learning queue. 
Analog to the prediction queue the processing (trainer) also has its queue and a corresponding process (Fig. \ref{fig:ga3c-block}. 
Figure \ref{fig:architectur} shows the architecture of the network).

The inputs of the network are the last four laser scans with 1081 values each and the orientation to the goal according to a compass value.
On a real robot the orientation can be calculated with a compass and the wheel encoder odometry.
Of course, a map can also be used if it is available.
The orientation is mapped from a value to a vector of size 128 ([0-360] $\rightarrow$ \{0,0,...,0,1,0,..,0\}) and 
is decoded as a one-hot vector, i.e. only the value for the direction to the goal has a one whereas all other vector elements are zero. 
The laser input is given in polar coordinates i.e. distance values are normalized between [0,1] with a maximum distance of 20 meter (Fig. \ref{fig:input-coding}). 

The next two layers of the network are 1D convolutions (1x9x16,1x5x32) with a stride of four and two.
We also tested 2D convolutions, but the overall performance was not better.
All filters are merged into a dense layer of 256 artificial neurons. 
The policy layer (V(s) or $\pi$) as well as the Q-value layer, which are the actor and the critic, use the dense layer to calculate the Q-value and policy value. 
The policy value is one-hot encoded and interpreted as a discrete action. 
An action is a pair of two values consisting of the angular and linear velocity. 
Since our robot is relative slow (max velocity of 0.6 m/s) we only use seven different actions from forward (0$^\circ$/s,0.6m/s) to turning left/right (+/-0.9*180$^\circ$/s, 0.2m/s). 
So, compared to other convolutional neural networks, e.g. for image classification, our network is flat.

Even more challenging is the design of an appropriate reward function. 
The reward function is part of the environment and evaluates how well agents calculate their actions (policy). 
It also gives the value function (Q(s,a)) which values the calculated action (s: state, a:action). 
In other words, it appropriately rewards each decision and future states - that is the transition - to evaluate the V(s) and Q(s,a) according to the Bellman equation. 
$$ 
Q(s_{t},a_{t}) \leftarrow (1-\alpha) \cdot \underbrace{Q(s_{t},a_{t})}_{\rm old~value} + ~~~~~~~~~~~~~~ 
$$ 
$$ 
\underbrace{\alpha}_{\rm learning~rate} \cdot  \overbrace{\bigg( \underbrace{r_{t}}_{\rm reward} + \underbrace{\gamma}_{\rm discount~factor} \cdot \underbrace{\max_{a}Q(s_{t+1}, a)}_{\rm estimate~of~optimal~future~value} \bigg) }^{\rm learned~value} 
$$ 
Now the question arises, if the network calculates an output (policy V(s)/$\pi()$, Q(s,a)) which return should be expected from the environment? 
Relatively simple are the rewards for the final (goal reached) and collision states that will be rewarded by 20 and -20. 
Since the initial reward value is zero, it leads to states Q(s,a) = 0 except for those that lead to the collision or reaching the goal. 
So any selected action, that does not cause a collision is considered a ''not bad'' action which leads to robots that circle around themselves
but do not try to reach the goal position. 
The probability of choosing good sequences of actions to reach a goal position, for long sequences of observations and actions along the robot path, tends to zero.  
So, we have to add additional intermediate rewards to distinguish non-collision states. 
Therefore, we define how the states and decisions that lead to a movement of the robot are rewarded. 
Our intermediate rewards consist of two terms and take into account how much the movement of the robot has taken the robot closer to the target position. 
\begin{enumerate} 
\item If the distance to the goal is smaller than before the robot gets a small positive reward otherwise a small negative reward. 
\item If the orientation of the robot is closer to the direction of the goal, it gets a small positive reward otherwise a small negative reward. 
\end{enumerate} 
Furthermore, we limit the maximum number of iterations per episode to 1000 cycles, i.e. if the robot has not reached the goal, its agent gets a negative reward. 
The update frequency of the robot is set to 2 Hz.  
The extended reward function leads to robust and fast learning results. 
While learning, the agents often reach circling states but are able to recover. 


\section{Training and Evaluation} 
\label{training} 

As already mentioned, we use several parallel simulation instances to train the agents. Figure \ref{fig:simulation3} shows a snapshot. 
There are three different environments with different degrees of difficulty, with the right one beeing a map of real environment of our University Lab. 
To speed up the learning, we train 32 different environment instances in parallel and in our example with eight trainers (Fig. \ref{fig:ga3c-block}).  

\begin{figure*}[!t] 
\centering 
\includegraphics[width=0.94\textwidth]{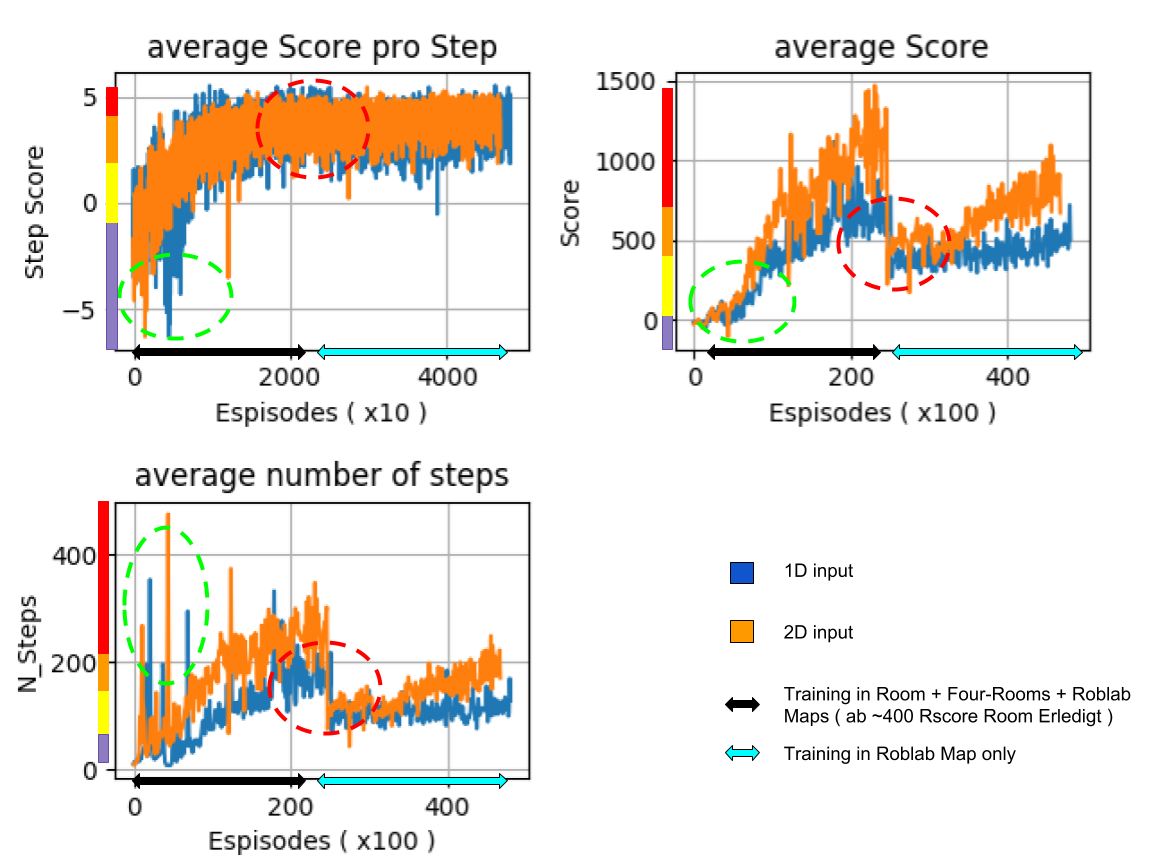} 
\caption{Training results for a 1D and 2D convolutional network for 20.000 episodes. Green circles: Circulating robots. Red circles: Reduction of the simulation to one environmental map (Lab).} 
\label{fig:results} 
\end{figure*}

Figure \ref{fig:results} shows the learning results for different input convolutions of the first two layers. 
The training time for 20.000 episodes was 25 minutes (1D convolution) vs. 7 hours (2D convolution) at an Intel core i7 and an Nvidia GeForce 1070ti. 
The overall performance of the networks is comparable.  
Green circles mark circling robots since the training episodes have a high number of steps per batch, i.e. long driving periods and low average score without reaching the goal position. 

The number of different environments is reduced to the map of our Lab after 20.000 epochs (red circle). The average score of 1500 means a collision free navigation in 95\% of the runs (75\% at 1000). At the end of the training all collision occure only in the lab environment of figure \ref{fig:simulation3} (left). This environment is build with hector slam based only on the 2D laser (without the 3D sensor) which leads to all of the small bins. These bins are the poles of the chairs. The chairs are larger obstacles on the real robot with the described combination of the 2D laser scanner and the RGBD camera.

\begin{figure}[!t] 
\centering 
\includegraphics[width=0.48\textwidth]{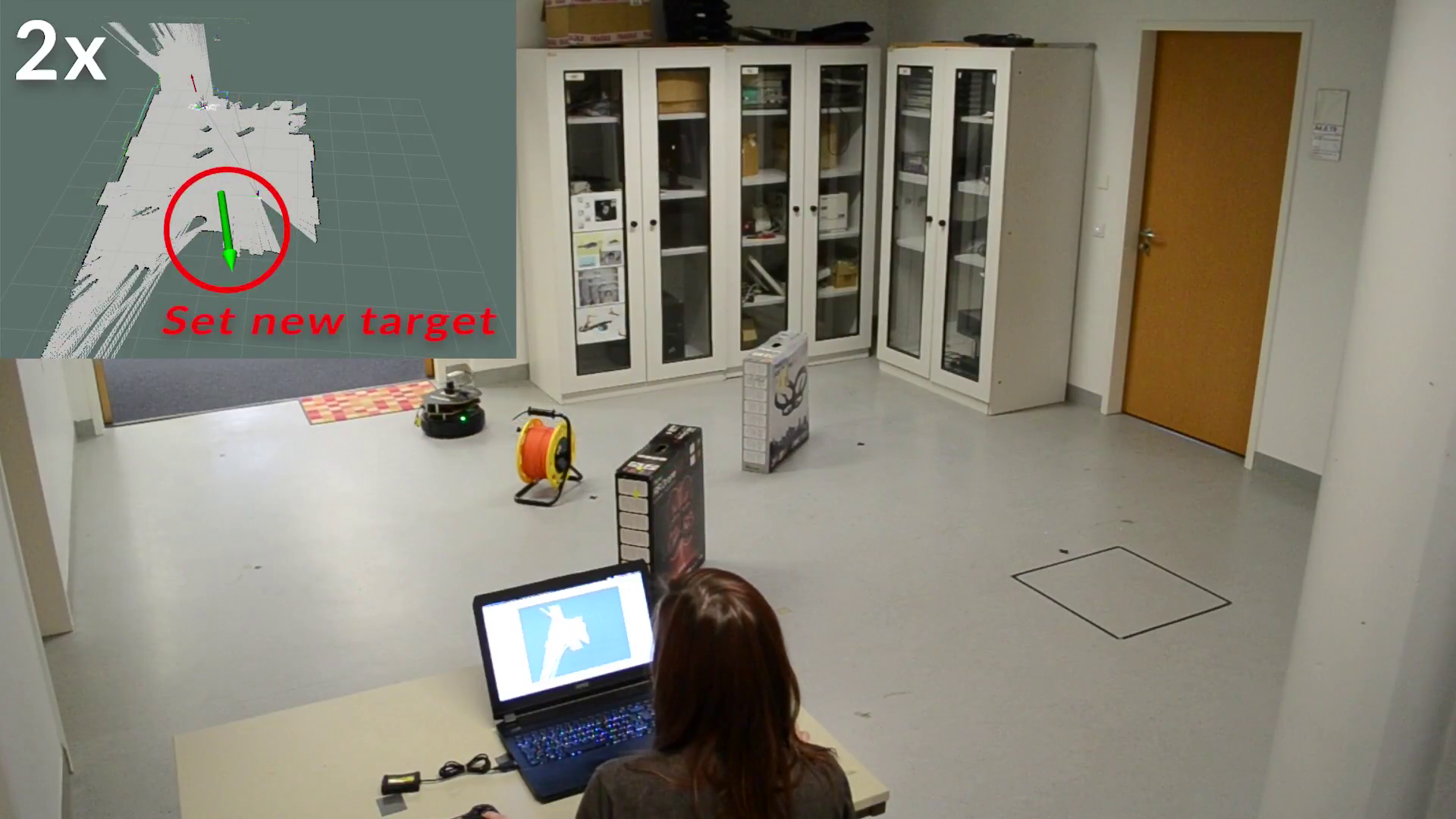} 
\caption{Online learning and testing on the robot in real environments. A new target is set with rviz and sent to the robot. The map is built with hector slam and used to visualize the position and to calculate the reward.} 
\label{fig:robot-eval} 
\end{figure}

Besides the offline learning, evaluation and online learning were also done with a real robot. 
Here the goal position is set at a laptop and sent to the robot (Fig. \ref{fig:robot-eval}). The laptop is also used to monitor the progress via rviz. 
A map is built with hector slam and used to visualize the position and to calculate the rewards. 
In our case, the offline learning was so good that we could not improve it by the online learning.
Furthermore, we observe a better obstacle avoidance in the real world than during offline learning. 
A possible explanation could be that the agent choses random actions during the learning to improve the performance, 
while this is not done during the real world tests.  

During our real world experiments we only observe one type of collision. The robot touches an obstacle with its back side. The type of obstacle is lower than the scan plane of the 2D laser scanner and only visible with the 3D camera. This type of obstacles disappeared in the input vector of the neural network when the distance of the robot and the obstacle is lower 25 cm because the Real Sense minimum depth distance is 25 cm. The neural network moves the robot back to its goal orientation when the obstacle disappeared and touch it with its back side. Presumably, a better 3D-RGBD sensor or a recurrent neural network with a longer training could avoid this problem.

\section{Conclusion} 
\label{sec:conclusion}

Data Preprocessing is often said by multitudes of people to be the most important part of the Machine Learning Algorithm.
Deep reinforcement learning approaches need a huge amount of positive and negative training examples to learn unsupervised. So, the main contribution of this paper is probably the integration of a massive parallel environment simulation for mobile robots into the deep reinforcement learning process. This leads to an infinte number of good training examples which improves generalization, avoids overfitting and overcomes the training example bottleneck. It leads us to a self-learned deep neural network with an incredible good performance (similar to human) in the real world. The simulator is about 1000 times faster than a unity, gazebo or player/stage simulator. Furthermore, it integrates different environments with different complexities in the learning process which boosts the learning process and robustness of our neural network. It reduces the learning process from years and months to hours. The code can be downloaded at GitHub.
The second contribution of this paper is the transfer from simulation to real-world environments. This is achieved by using the Nvidia Jetson board for the inferencing of the neural network and by combining a 2D laser scanner with a 3D-RGBD camera to generate appropriated network inputs. Since 2D laser scanner a widly used on mobile robots our approach can be used on all of these robots. 

Needless to say a lot of work remains to be done. More quantitative evaluation, different network architectures (DDPG, LSTMs, …) and the behavior of multiple robots / swarm robotics have to (and can!) be examined in environments of different complexities. Stay tuned, updates will folow. 

%

\bibliographystyle{plainnat}
\bibliography{literatur}

\begin{thebibliography}{14}
\providecommand{\natexlab}[1]{#1}
\providecommand{\url}[1]{\texttt{#1}}
\expandafter\ifx\csname urlstyle\endcsname\relax
  \providecommand{\doi}[1]{doi: #1}\else
  \providecommand{\doi}{doi: \begingroup \urlstyle{rm}\Url}\fi

\bibitem[Babaeizadeh et~al.(2016)Babaeizadeh, Frosio, Tyree, Clemons, and
  Kautz]{DBLP:journals/corr/BabaeizadehFTCK16}
Mohammad Babaeizadeh, Iuri Frosio, Stephen Tyree, Jason Clemons, and Jan Kautz.
\newblock {GA3C:} gpu-based {A3C} for deep reinforcement learning.
\newblock \emph{CoRR}, abs/1611.06256, 2016.
\newblock URL \url{http://arxiv.org/abs/1611.06256}.

\bibitem[Chen et~al.(2017)Chen, Everett, Liu, and
  How]{DBLP:journals/corr/ChenELH17}
Yu~Fan Chen, Michael Everett, Miao Liu, and Jonathan~P. How.
\newblock Socially aware motion planning with deep reinforcement learning.
\newblock \emph{CoRR}, abs/1703.08862, 2017.
\newblock URL \url{http://arxiv.org/abs/1703.08862}.

\bibitem[{Echeverria} et~al.(2011){Echeverria}, {Lassabe}, {Degroote}, and
  {Lemaignan}]{5980252}
G.~{Echeverria}, N.~{Lassabe}, A.~{Degroote}, and S.~{Lemaignan}.
\newblock Modular open robots simulation engine: Morse.
\newblock In \emph{2011 IEEE International Conference on Robotics and
  Automation}, pages 46--51, May 2011.
\newblock \doi{10.1109/ICRA.2011.5980252}.

\bibitem[{Ko} et~al.(2017){Ko}, {Hong}, {Kim}, {Kwon}, and {Yoo}]{7881744}
B.~{Ko}, H.J.~{Choi and }C. {Hong}, J.H. {Kim}, O.C. {Kwon}, and C.~D. {Yoo}.
\newblock Neural network-based autonomous navigation for a homecare mobile
  robot.
\newblock In \emph{2017 IEEE International Conference on Big Data and Smart
  Computing (BigComp)}, pages 403--406, Feb 2017.
\newblock \doi{10.1109/BIGCOMP.2017.7881744}.

\bibitem[{Koenig} and {Howard}(2004)]{1389727}
N.~{Koenig} and A.~{Howard}.
\newblock Design and use paradigms for gazebo, an open-source multi-robot
  simulator.
\newblock In \emph{2004 IEEE/RSJ International Conference on Intelligent Robots
  and Systems (IROS) (IEEE Cat. No.04CH37566)}, volume~3, pages 2149--2154
  vol.3, Sep. 2004.
\newblock \doi{10.1109/IROS.2004.1389727}.

\bibitem[Kohlbrecher et~al.(2011)Kohlbrecher, Meyer, von Stryk, and
  Klingauf]{KohlbrecherMeyerStrykKlingaufFlexibleSlamSystem2011}
S.~Kohlbrecher, J.~Meyer, O.~von Stryk, and U.~Klingauf.
\newblock A flexible and scalable slam system with full 3d motion estimation.
\newblock In \emph{Proc. IEEE International Symposium on Safety, Security and
  Rescue Robotics (SSRR)}. IEEE, November 2011.

\bibitem[Lillicrap et~al.(2015)Lillicrap, Hunt, Pritzel, Heess, Erez, Tassa,
  Silver, and Wierstra]{DBLP:journals/corr/LillicrapHPHETS15}
Timothy~P. Lillicrap, Jonathan~J. Hunt, Alexander Pritzel, Nicolas Heess, Tom
  Erez, Yuval Tassa, David Silver, and Daan Wierstra.
\newblock Continuous control with deep reinforcement learning.
\newblock \emph{CoRR}, abs/1509.02971, 2015.
\newblock URL \url{http://arxiv.org/abs/1509.02971}.

\bibitem[Mnih et~al.(2013)Mnih, Kavukcuoglu, Silver, Graves, Antonoglou,
  Wierstra, and Riedmiller]{DBLP:journals/corr/MnihKSGAWR13}
Volodymyr Mnih, Koray Kavukcuoglu, David Silver, Alex Graves, Ioannis
  Antonoglou, Daan Wierstra, and Martin~A. Riedmiller.
\newblock Playing atari with deep reinforcement learning.
\newblock \emph{CoRR}, abs/1312.5602, 2013.
\newblock URL \url{http://arxiv.org/abs/1312.5602}.

\bibitem[Mnih et~al.(2016)Mnih, Badia, Mirza, Graves, Lillicrap, Harley,
  Silver, and Kavukcuoglu]{DBLP:journals/corr/MnihBMGLHSK16}
Volodymyr Mnih, Adri{\`{a}}~Puigdom{\`{e}}nech Badia, Mehdi Mirza, Alex Graves,
  Timothy~P. Lillicrap, Tim Harley, David Silver, and Koray Kavukcuoglu.
\newblock Asynchronous methods for deep reinforcement learning.
\newblock \emph{CoRR}, abs/1602.01783, 2016.
\newblock URL \url{http://arxiv.org/abs/1602.01783}.

\bibitem[{Rohmer} et~al.(2013){Rohmer}, {Singh}, and {Freese}]{6696520}
E.~{Rohmer}, S.~P.~N. {Singh}, and M.~{Freese}.
\newblock V-rep: A versatile and scalable robot simulation framework.
\newblock In \emph{2013 IEEE/RSJ International Conference on Intelligent Robots
  and Systems}, pages 1321--1326, Nov 2013.
\newblock \doi{10.1109/IROS.2013.6696520}.

\bibitem[Shabbir and Anwer(2018)]{DBLP:journals/corr/abs-1803-07608}
Jahanzaib Shabbir and Tarique Anwer.
\newblock A survey of deep learning techniques for mobile robot applications.
\newblock \emph{CoRR}, abs/1803.07608, 2018.
\newblock URL \url{http://arxiv.org/abs/1803.07608}.

\bibitem[Tai et~al.(2017)Tai, Paolo, and Liu]{DBLP:journals/corr/TaiPL17}
Lei Tai, Giuseppe Paolo, and Ming Liu.
\newblock Virtual-to-real deep reinforcement learning: Continuous control of
  mobile robots for mapless navigation.
\newblock \emph{CoRR}, abs/1703.00420, 2017.
\newblock URL \url{http://arxiv.org/abs/1703.00420}.

\bibitem[Tan et~al.(2018)Tan, Zhang, Coumans, Iscen, Bai, Hafner, Bohez, and
  Vanhoucke]{DBLP:journals/corr/abs-1804-10332}
Jie Tan, Tingnan Zhang, Erwin Coumans, Atil Iscen, Yunfei Bai, Danijar Hafner,
  Steven Bohez, and Vincent Vanhoucke.
\newblock Sim-to-real: Learning agile locomotion for quadruped robots.
\newblock \emph{CoRR}, abs/1804.10332, 2018.
\newblock URL \url{http://arxiv.org/abs/1804.10332}.

\bibitem[Wang et~al.(2016)Wang, Bapst, Heess, Mnih, Munos, Kavukcuoglu, and
  de~Freitas]{DBLP:journals/corr/WangBHMMKF16}
Ziyu Wang, Victor Bapst, Nicolas Heess, Volodymyr Mnih, R{\'{e}}mi Munos, Koray
  Kavukcuoglu, and Nando de~Freitas.
\newblock Sample efficient actor-critic with experience replay.
\newblock \emph{CoRR}, abs/1611.01224, 2016.
\newblock URL \url{http://arxiv.org/abs/1611.01224}.

\end{thebibliography}

\end{document}